# Causal inference for climate change events from satellite image time series using computer vision and deep learning

Vikas Ramachandra

Adjunct Professor, Data Science

University of California, Berkeley

Berkeley, CA

virama@berkeley.edu

## Abstract

We propose a method for causal inference using satellite image time series, in order to determine the treatment effects of interventions which impact climate change, such as deforestation. Simply put, the aim is to quantify the 'before versus after' effect of climate related human driven interventions, such as urbanization; as well as natural disasters, such as hurricanes and forest fires. As a concrete example, we focus on quantifying forest tree cover change/ deforestation due to human led causes. The proposed method involves the following steps. First, we uae computer vision and machine learning/deep learning techniques to detect and quantify forest tree coverage levels over time, at every time epoch. We then look at this time series to identify changepoints. Next, we estimate the expected (forest tree cover) values using a Bayesian structural causal model and projecting/forecasting the counterfactual. This is compared to the values actually observed post intervention, and the difference in the two values gives us the effect of the intervention (as compared to the non intervention scenario, i.e. what would have possibly happened without the intervention). As a specific use case, we analyze deforestation levels before and after the hyperinflation event (intervention) in Brazil (which ended in 1993-94), for the Amazon rainforest region, around Rondonia, Brazil. For this deforestation use case, using our causal inference framework can help causally attribute *change/reduction* in forest tree cover and increasing deforestation rates due to human activities at various points in time.

## 1. Introduction and background

Over the last two decades, deforestation has become a major problem. Various institutions and organisations are trying to monitor land use and protect forests by using large amount of satellite imagery, leveraging the remote sensing technology. Satellite remote sensing allows to observe the changes and calculate the deforestation rate. Based on these images, certain algorithm/methodology can be implemented on the top of it to build an automated forest cover detection system. We will start with the image processing which will act as our baseline method, and also use deep learning based segmentation methods to detect forest cover. The output of this algorithm is then fed to changepoint detection and a temporal causal inference technique. This enables us to quantify the effect on deforestation rate due to human led interventions. This is done by computing the forest cover before, forecasting the counterfactual, and comparing it to the forest cover after an intervention.

This paper is organized as follows. Section 2 provides details about the dataset used, followed by Section 3, which provides details about each step in the algorithm (image processing, computer vision and deep learning for forest cover detection, changepoint analysis, and time series causal inference algorithms). Concluding remarks are provided in Section 4.

## 2. Details of the dataset

For our analysis, we use the Google Earth satellite image dataset. The region selected was in the Amazonian rainforest in Brazil, specifically around Rondonia, Brazil. Satellite images of this region were gathered, from 1984-2018. Due to frontier expansion, it has been noted that the forest tree cover has been steadily declining in this region over time [4]. Brazil also saw a period of hyperinflation which affected economic growth and expansion activities [3] For the specific intervention or event, we consider the end of hyperinflation around 1993-94, and quantify its effect on deforestation. The hypothesis is that the end of hyperinflation fuelled rapid growth and led to even higher rates of deforestation, than the period before and during the hyperinflation. In the next sections, we discuss the algorithm steps used to verify this hypothesis.

## 3. Proposed algorithm details

The proposed algorithm consists of the following steps:
   A. Computer vision and deep learning (CV/DL) based image feature extraction, segmentation and quantification of forest tree cover at each time epoch.
   B. Changepoint detection
   C. Time series causal inference for CV/DL image features

<u>A. Image processing/ computer vision and deep learning for feature extraction and segmentation of forest cover</u>

The aim of this step is to detect tree cover and aggregating image pixels for regions occupied by trees, in the satellite images.

## Method A.1: HSV masking

By extracting the hue, saturation and intensity i.e. HSV channels from an image we are essentially performing color isolation. This will enable better thresholding and selection of regions of interest. We will convert the RGB image to HSV because HSV format separates the image intensity from the chroma or color information unlike RGB which could not separate the color intensity from color information.

In the HSV space, H is hue that determines the color you want. S is saturation that determines how intense the color is. V is value that determines the lightness of the image. We will mask the forest cover area using this method of HSV masking. In order to determine the upper and lower HSV range respectively for the forest cover, we created an interactive widget which lets the user select the ranges for these parameters on a training image subset. The selected values are then applied as a threshold across all images.

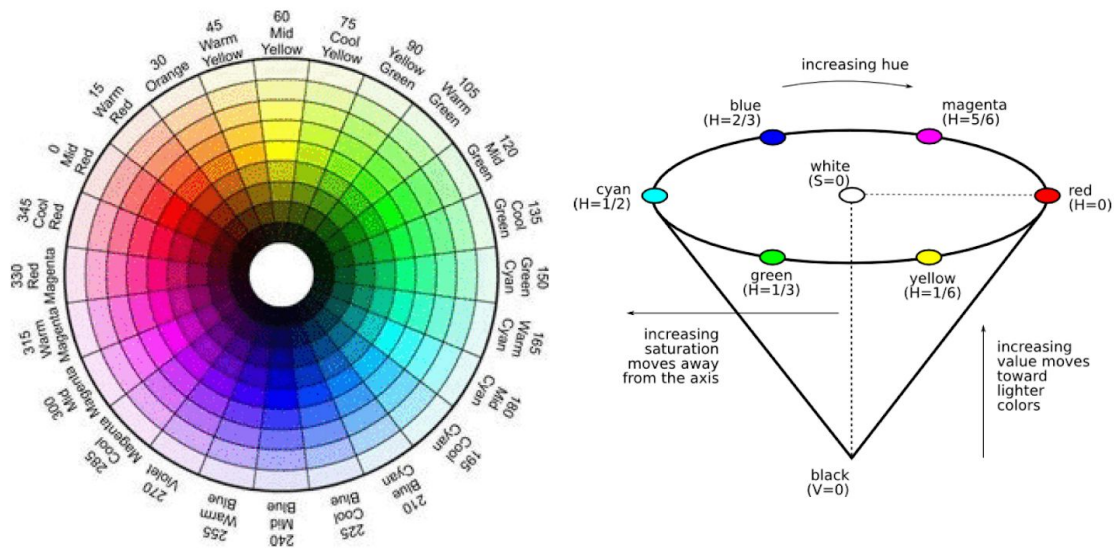

Fig. HSV Representation of Color - image from nmt.edu

Note that if the image is noisy then this technique needs to be combined with denoising techniques such as a Gaussian blur filtering operation on the image. Another viable approach is to use K-means clustering for color segmentation.

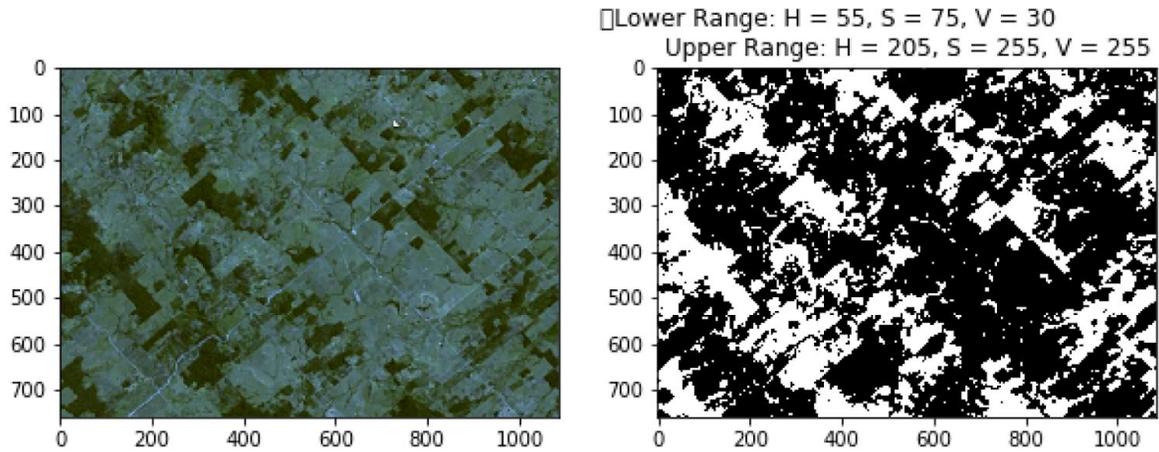

Figure 1: Left: Original input image. Right: Binary mask with with white=forest, and black=other land

The original input images are 8 bit (0-255) and 3 channel (RGB). Lower and upper ranges of HSV for forest cover are found to be [55, 75, 30] and [205, 255, 255] respectively, based on user inputs via the interactive widget. For generating the mask, if the pixel HSV values lie in between these two bounds, we assign a mask pixel value to 255 (white) else to 0 (black). Here, as shown in figure 1 above, the white region is the object we have to mask(here the forest cover) and black region is the background(land). After creating the mask, the algorithm counts the.number of white pixels and divides it by the total number of pixels. This will give us the fraction of forest cover detected in the image, as shown in Figure 2 below.

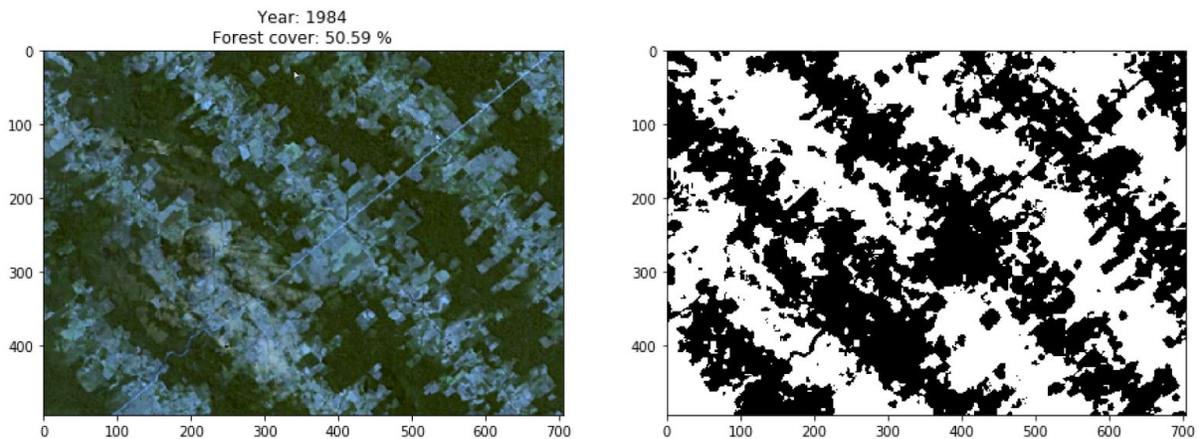

Figure 2: Left: Original image. Right: binary mask for forest cover, Top: Percentage of forest cover detected (50.59% for the year 1984)

**Method A.2: Deep learning based semantic segmentation**

As an alternative to the above method, we train a deep learning framework for semantic segmentation, which classifies every pixel of the image into one of the two classes (forest cover or not). For this task classes will be forest and background. The general semantic segmentation model is a deep neural network architecture which can be thought of as an encoder network followed by a decoder network.

Encoder Network:
- Generally consist of pre-trained classification network like VGG16/VGG19/ResNet/DenseNet.
- It takes an input image and generates a high-dimensional feature vector.
- Aggregates features at multiple levels.

Decoder Network:
- It takes a high dimensional feature vector and generates a semantic segmentation mask
- Decodes features aggregated by encoder at multiple levels

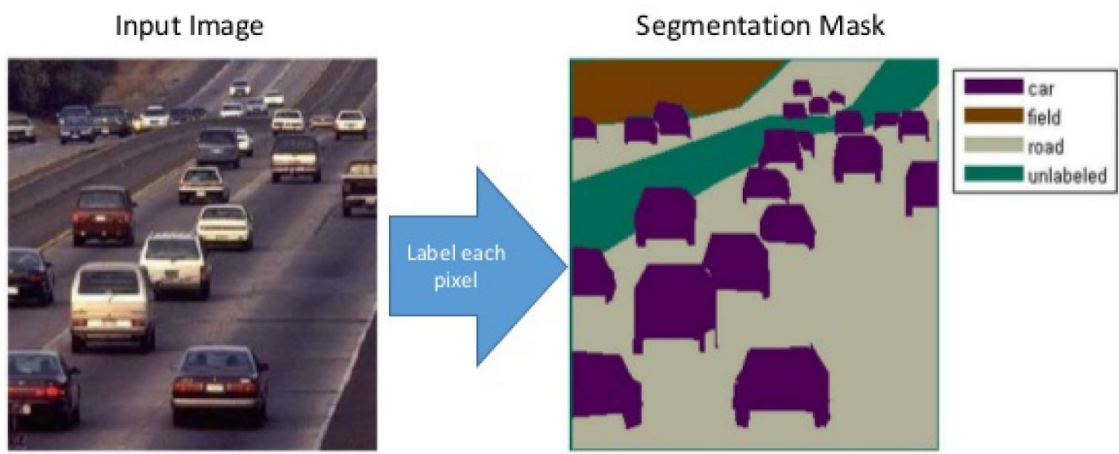

Figure 3: Illustration of the output of existing semantic segmentation algorithms applied to traffic images

For our implementation we used the U-Net [7], a popular semantic segmentation model, which consists of convolution/downsampling layers, max. pooling, followed by deconvolution/upsampling layers.

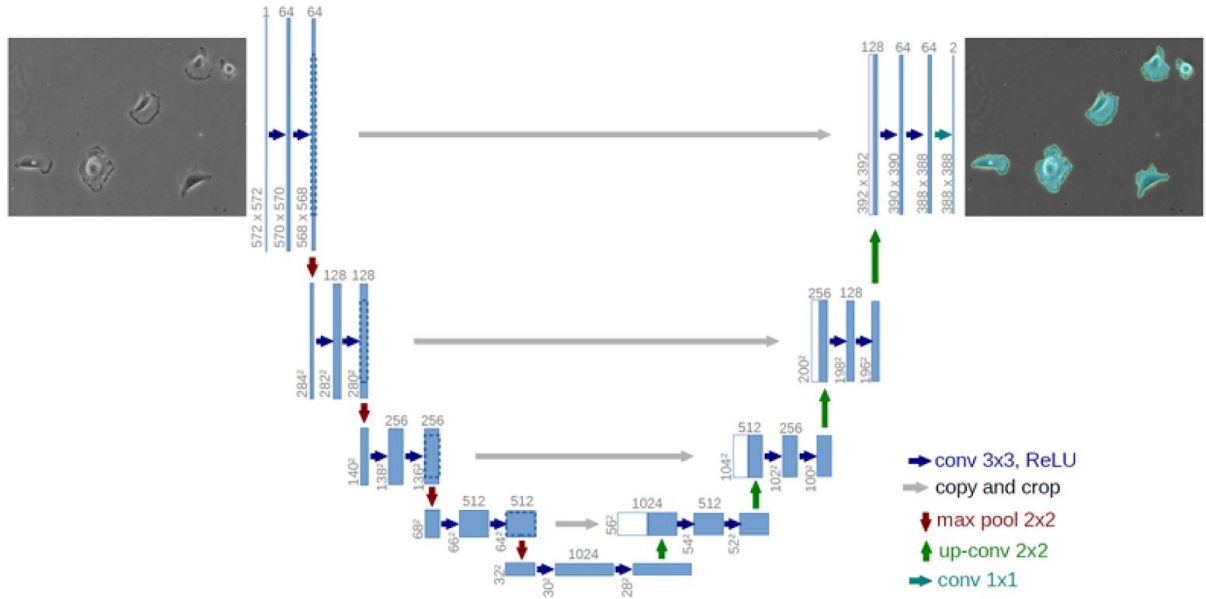

Figure 4: U-Net architecture

We labeled the dataset using the Labelme [6] annotation tool to create ground truth segmentation masks for multiple images in the training set, which is used for training the model. Testing data contains the images of the same target area that we used in Image Processing, for comparison. We took 10 images for ground truth labeling as well as training (other than the 1984-2018 yearly images) and 35 images of the target region from 1984 to 2018 for testing. During training, the Jaccard score [5] was used as the evaluation metric, and 5 fold cross validation was applied. Data augmentation was performed using image rotation and flips. Adam optimizer was used with binary cross entropy loss. The network losses and Jaccard indices are plotted below in figure 4a against epochs. The network parameters and layers are as shown in figure 5. Some example results of the semantic segmentation from this U-Net is shown in figure 6.

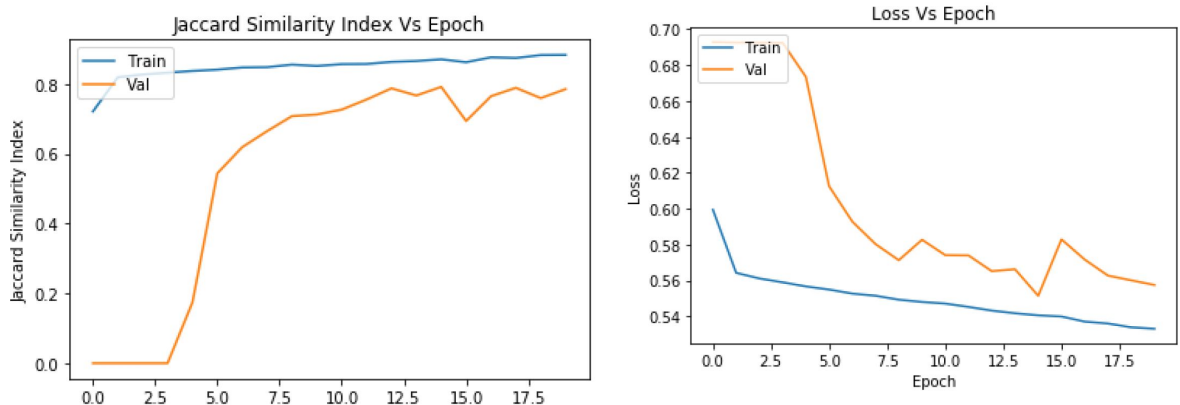

Figure 4a: Jaccard index and binary cross entropy loss during training (Versus epochs)

```
----------------------------------------------------------------
        Layer (type)               Output Shape         Param #
================================================================
            Conv2d-1        [-1, 32, 128, 128]             896
       BatchNorm2d-2        [-1, 32, 128, 128]              64
         LeakyReLU-3        [-1, 32, 128, 128]               0
        conv_block-4        [-1, 32, 128, 128]               0
            Conv2d-5        [-1, 32, 128, 128]           9,248
       BatchNorm2d-6        [-1, 32, 128, 128]              64
         LeakyReLU-7        [-1, 32, 128, 128]               0
        conv_block-8        [-1, 32, 128, 128]               0
            Conv2d-9          [-1, 64, 64, 64]          18,496
      BatchNorm2d-10          [-1, 64, 64, 64]             128
        LeakyReLU-11          [-1, 64, 64, 64]               0
       conv_block-12          [-1, 64, 64, 64]               0
           Conv2d-13          [-1, 64, 64, 64]          36,928
      BatchNorm2d-14          [-1, 64, 64, 64]             128
        LeakyReLU-15          [-1, 64, 64, 64]               0
       conv_block-16          [-1, 64, 64, 64]               0
           Conv2d-17         [-1, 128, 32, 32]          73,856
      BatchNorm2d-18         [-1, 128, 32, 32]             256
        LeakyReLU-19         [-1, 128, 32, 32]               0
       conv_block-20         [-1, 128, 32, 32]               0
           Conv2d-21         [-1, 128, 32, 32]         147,584
      BatchNorm2d-22         [-1, 128, 32, 32]             256
        LeakyReLU-23         [-1, 128, 32, 32]               0
       conv_block-24         [-1, 128, 32, 32]               0
           Conv2d-25         [-1, 128, 16, 16]         147,584
      BatchNorm2d-26         [-1, 128, 16, 16]             256
        LeakyReLU-27         [-1, 128, 16, 16]               0
       conv_block-28         [-1, 128, 16, 16]               0
           Conv2d-29         [-1, 256, 32, 32]         590,080
      BatchNorm2d-30         [-1, 256, 32, 32]             512
        LeakyReLU-31         [-1, 256, 32, 32]               0
       conv_block-32         [-1, 256, 32, 32]               0
           Conv2d-33          [-1, 64, 32, 32]         147,520
      BatchNorm2d-34          [-1, 64, 32, 32]             128
        LeakyReLU-35          [-1, 64, 32, 32]               0
       conv_block-36          [-1, 64, 32, 32]               0
           Conv2d-37         [-1, 128, 64, 64]         147,584
      BatchNorm2d-38         [-1, 128, 64, 64]             256
        LeakyReLU-39         [-1, 128, 64, 64]               0
       conv_block-40         [-1, 128, 64, 64]               0
           Conv2d-41          [-1, 32, 64, 64]          36,896
      BatchNorm2d-42          [-1, 32, 64, 64]              64
        LeakyReLU-43          [-1, 32, 64, 64]               0
       conv_block-44          [-1, 32, 64, 64]               0
           Conv2d-45        [-1, 64, 128, 128]          36,928
      BatchNorm2d-46        [-1, 64, 128, 128]             128
        LeakyReLU-47        [-1, 64, 128, 128]               0
       conv_block-48        [-1, 64, 128, 128]               0
```

```
        Conv2d-49            [-1, 1, 128, 128]             577
    BatchNorm2d-50            [-1, 1, 128, 128]               2
      LeakyReLU-51            [-1, 1, 128, 128]               0
     conv_block-52            [-1, 1, 128, 128]               0
================================================================
Total params: 1,396,419
Trainable params: 1,396,419
Non-trainable params: 0
----------------------------------------------------------------
Input size (MB): 0.19
Forward/backward pass size (MB): 119.50
Params size (MB): 5.33
Estimated Total Size (MB): 125.01
----------------------------------------------------------------
```

Figure 5: U-Net layers and parameters.

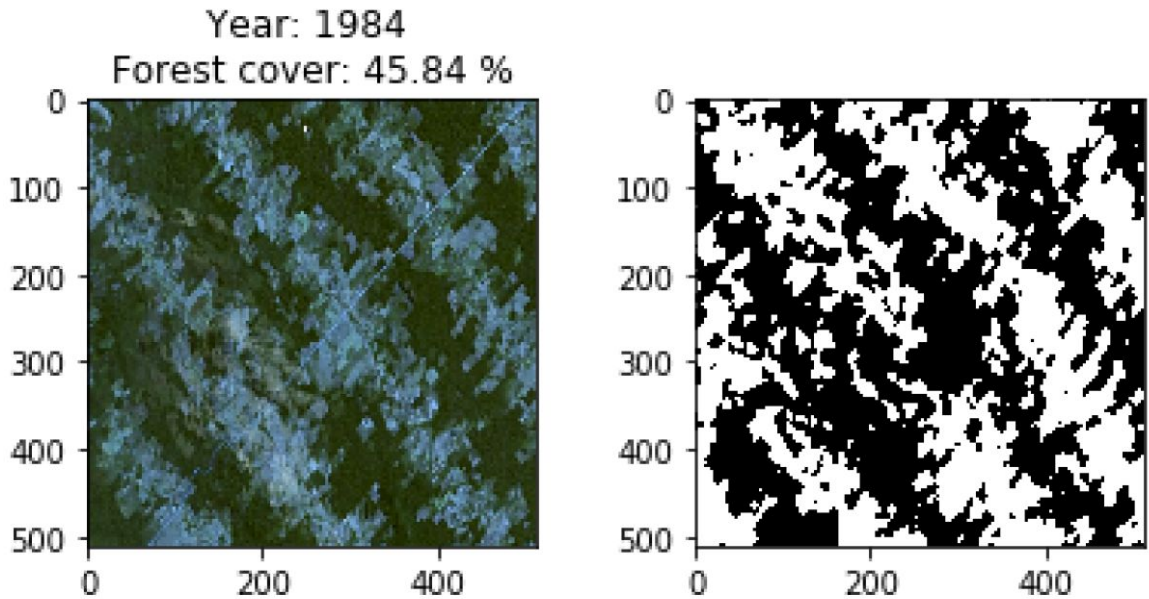

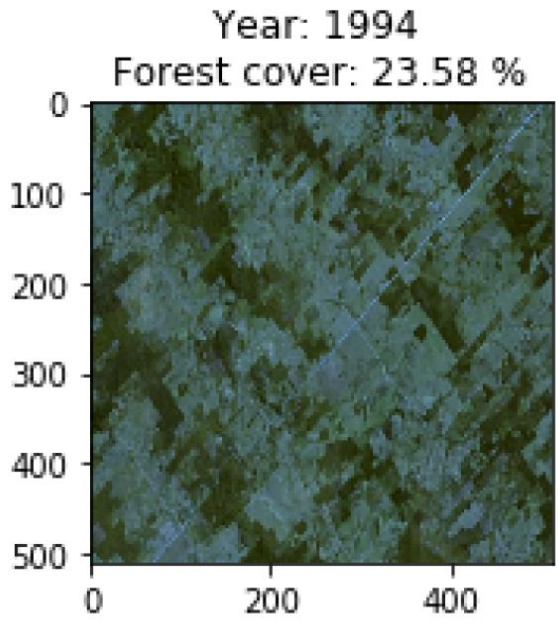
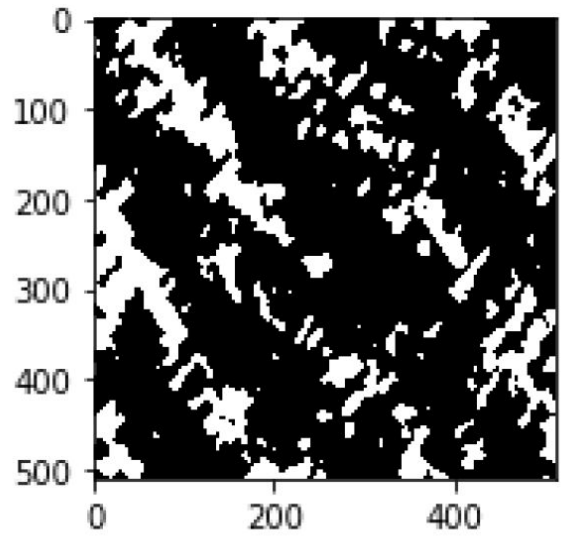
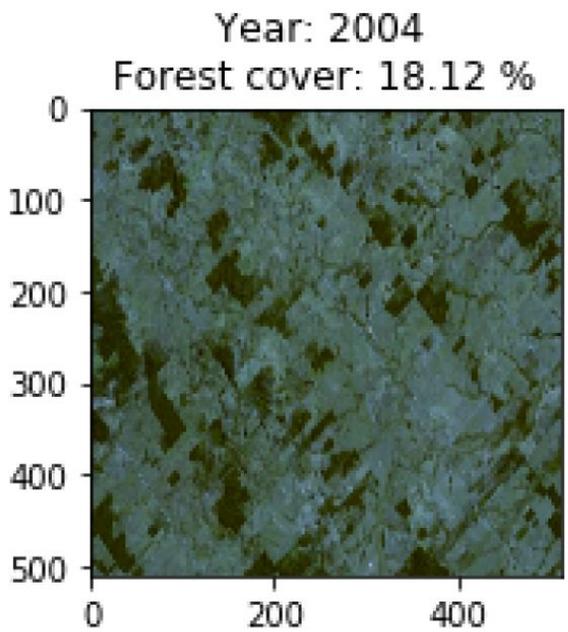
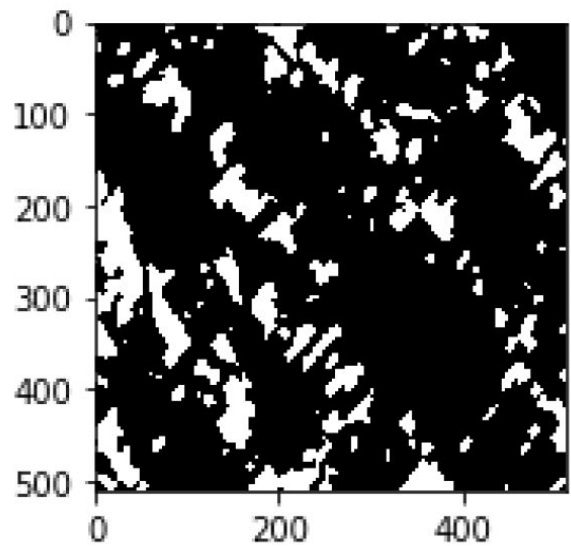

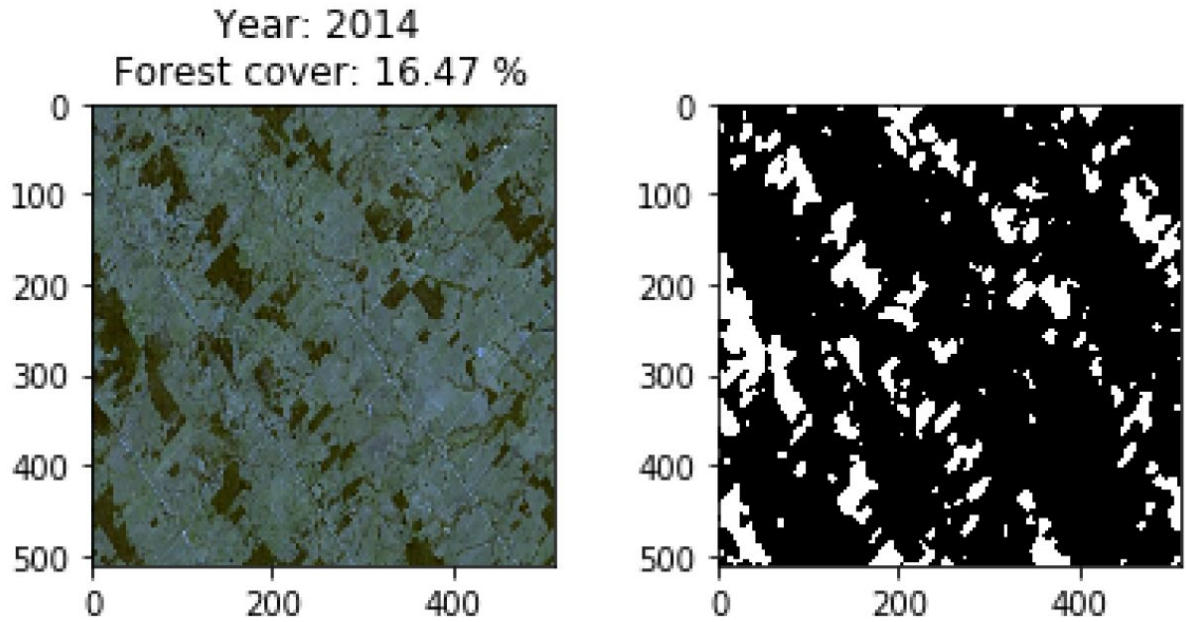

Figure 6: Results of the U-Net algorithm: some examples

**Time series construction**

For either method A.1 or A,2, the algorithm is run and forest cover calculations are done for each year's image, and the forest cover time series is constructed. Figure 7 below shows the forest cover time series plot.

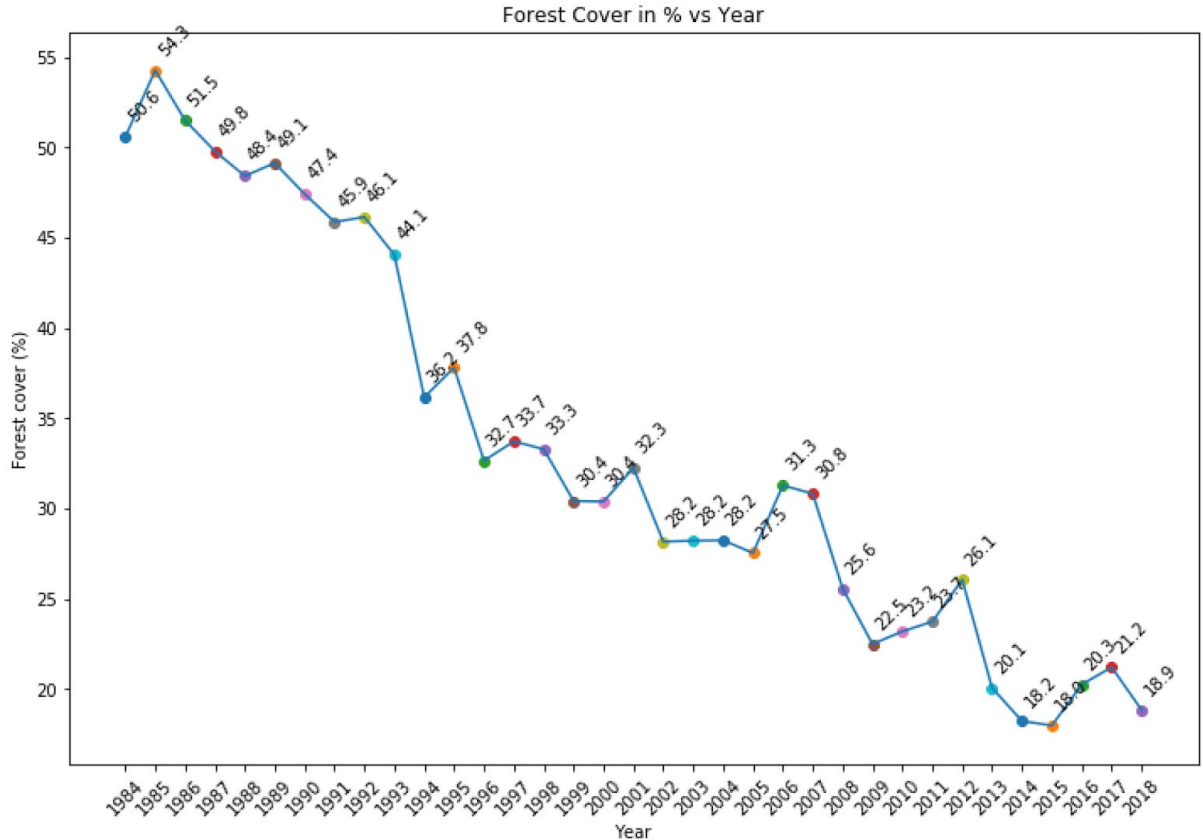

Figure 7: Forest tree cover percentage (Y axis) versus year (X axis) plot

We can see there is a gradual decrease in the forest cover over the years, pointing towards deforestation. Next, we use this time series for changepoint detection, and causal inference to compare before-after effects around the changepoint.

B.Changepoint detection

Our aim is to check whether the changepoint detection algorithm can identify the (end of hyperinflation) intervention year correctly from the above time series in Figure 7, thus verifying that the forest cover statistics before and after the event are indeed, different and the difference is statistically significant.

We use the mean and variance based changepoint detection algorithm in the changepoint R package; the reader can find more details in the related documentation [1]. We recap some of the details below.

Time series changepoint detection using the mean and variance can be described as follows. Changepoint detection is the name given to the problem of estimating the point at which the statistical properties of a sequence of observations change.

Let us assume we have an ordered sequence of data: $y_{1:n} = (y_1, \ldots, y_n)$. Changepoint occurs within this set when there exists a time: $\tau \in \{1, \ldots, n-1\}$, such that the statistical properties of $\{y_1, \ldots, y_\tau\}$ and $\{y_{\tau+1}, \ldots, y_n\}$ are different in some way.

**Single changepoint detection:**
We briefly recap the likelihood based framework for changepoint detection. Instead of considering the more general problem of identifying $\tau_{1:m}$ changepoint positions, we only consider the identification of a single changepoint here, for simplicity. The detection of a single changepoint can be posed as a hypothesis test. The null hypothesis, $H_0$, corresponds to no changepoint (m = 0) and the alternative hypothesis, $H_1$, is a single changepoint (m = 1). We now recap the general likelihood ratio based approach to test this hypothesis.

A test statistic can be constructed which we will use to decide whether a change has occurred. The likelihood ratio method requires the calculation of the maximum log-likelihood under both the null and alternative hypotheses. For the null hypothesis the maximum log-likelihood is log $p(y_{1:n}|\Theta^\wedge)$, where p(.) is the probability density function associated with the distribution of the data and $\Theta^\wedge$ is the maximum likelihood estimate of the parameters.

Under the alternative hypothesis, consider a model with a changepoint at $\tau_1$ with $\tau_1 \in \{1,2,\ldots, n-1\}$ Then the maximum log likelihood for a given $\tau_1$ is:

$$ML(\tau_1) = \log p(y_{1:\tau_1}|\hat{\theta}_1) + \log p(y_{(\tau_1+1):n}|\hat{\theta}_2).$$

Given the discrete nature of the changepoint location, the maximum log-likelihood value under the alternative is simply $\max_{\tau_1} ML(\tau_1)$, where the maximum is taken over all possible changepoint locations. The test statistic is thus

$$\lambda = 2\left[\max_{\tau_1} ML(\tau_1) - \log p(y_{1:n}|\hat{\theta})\right].$$

The test involves choosing a threshold, c, such that we reject the null hypothesis if $\lambda >$ c. If we reject the null hypothesis, i.e., detect a changepoint, then we estimate its position as $\tau_1$, the value of $\tau$ that maximizes $ML(\tau_1)$.

The test statistics used here are the mean and the variance.

As shown in figure 7 below, a changepoint is detected around the year 1994 (blue arrow), which is at the time the intervention happened (end of the hyperinflation) confirming the validity of the changepoint algorithm. Alternatively, an expert with domain knowledge could also pick the changepoint manually by visual inspection of the tree cover time series plot.

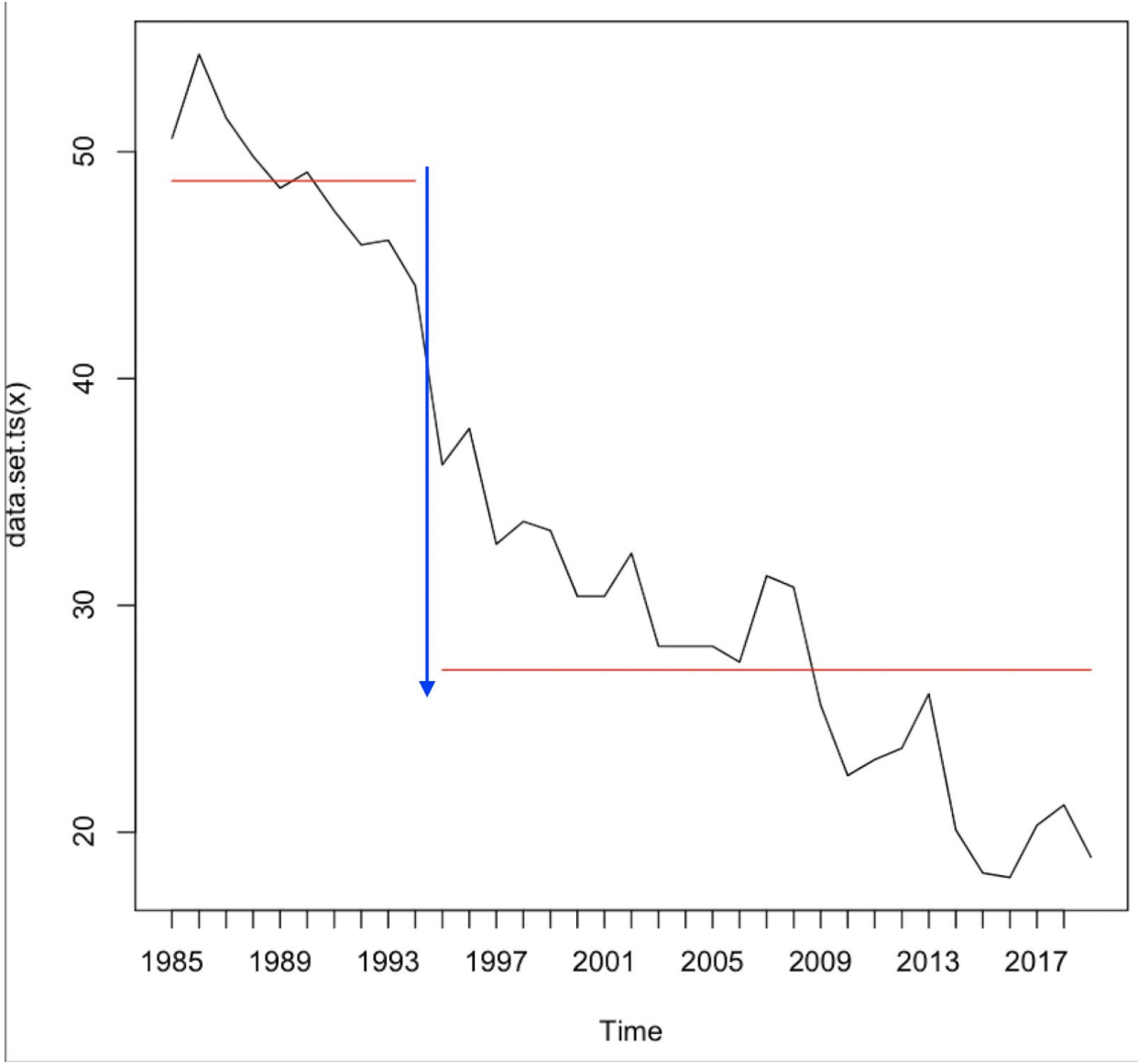

Figure 7: Changepoint detected using mean and variance at year 1994 for tree cover time series

C. Time series causal inference for CV/DL image features

For estimating treatment effect of the intervention, first we need to fit a model and forecast the counterfactual using the time series before the intervention and then compare that forecast to the actual time series recorded after the intervention. For this, we use the CausalImpact R package [2]. This package is based on causal inference for time series data using **Bayesian structural**

**models,** we recap some of the details below. The original paper has a much more comprehensive discussion about the methods used, and the reader is referred to it for more information [2].

The above method uses state space models and flexible Bayesian priors to fit a time series model pre-intervention, and forecast/predict the counterfactual based on the fit model.

Structural time-series models are state-space models for time-series data. They can be defined in terms of a pair of equations:

$$y_t = Z_t^T \alpha_t + \varepsilon_t,$$

$$\alpha_{t+1} = T_t \alpha_t + R_t \eta_t,$$

where $\varepsilon_t \sim N(0, \sigma_t^2)$ and $\eta_t \sim N(0, Q_t)$ are independent of all other unknowns. The first equation above is the observation equation; it links the observed data $y_t$ to a latent d-dimensional state vector $\alpha_t$. The second equation above is the state equation; it governs the evolution of the state vector $\alpha_t$ through time. Here, $y_t$ is a scalar observation, $Z_t$ is a d-dimensional output vector, $T_t$ is a $d \times d$ transition matrix, $R_t$ is a $d \times q$ control matrix, $\varepsilon_t$ is a scalar observation error with noise variance $\sigma_t$, and $\eta_t$ is a q-dimensional system error with a $q \times q$ state-diffusion matrix $Q_t$, where $q \leq d$.

The above parameters are learned using a Bayesian framework. Posterior inference is based on the Markov chain Monte Carlo (MCMC) technique, and a Gibbs sampler. Subtracting the predicted from the observed response during the post-intervention period gives a semiparametric Bayesian posterior distribution for the causal effect.

The input features are as derived in step A (tree cover percentage calculation). and described above. Formally, the covariates 'x' are location variables, and the output variable 'y' is the forest tree cover percentage at each given time point, and the treatment/ intervention variable is the location of the changepoint from step B.

The results of this step is shown in figure 8 below.

Top panel in figure 8 shows the counterfactual estimate (dashed line) and confidence intervals for it, and the solid line is the actual observed values (beyond 1994, the intervention event of hyperinflation completion). The middle panel in figure 8 Shows the estimated treatment effect, here the difference in the forest cover yearly between no intervention (projected counterfactual) and actual observed forest cover values. E.g.: at X=25, which is the year 2019, the tree cover difference is -25% (reduction) as compared to the projected no-intervention or 'before' values, this shows that the *end of hyperinflation event did cause an increased rate of deforestation and frontier expansion and hence, higher reduction in forest tree cover over the years, which validates our hypothesis.*

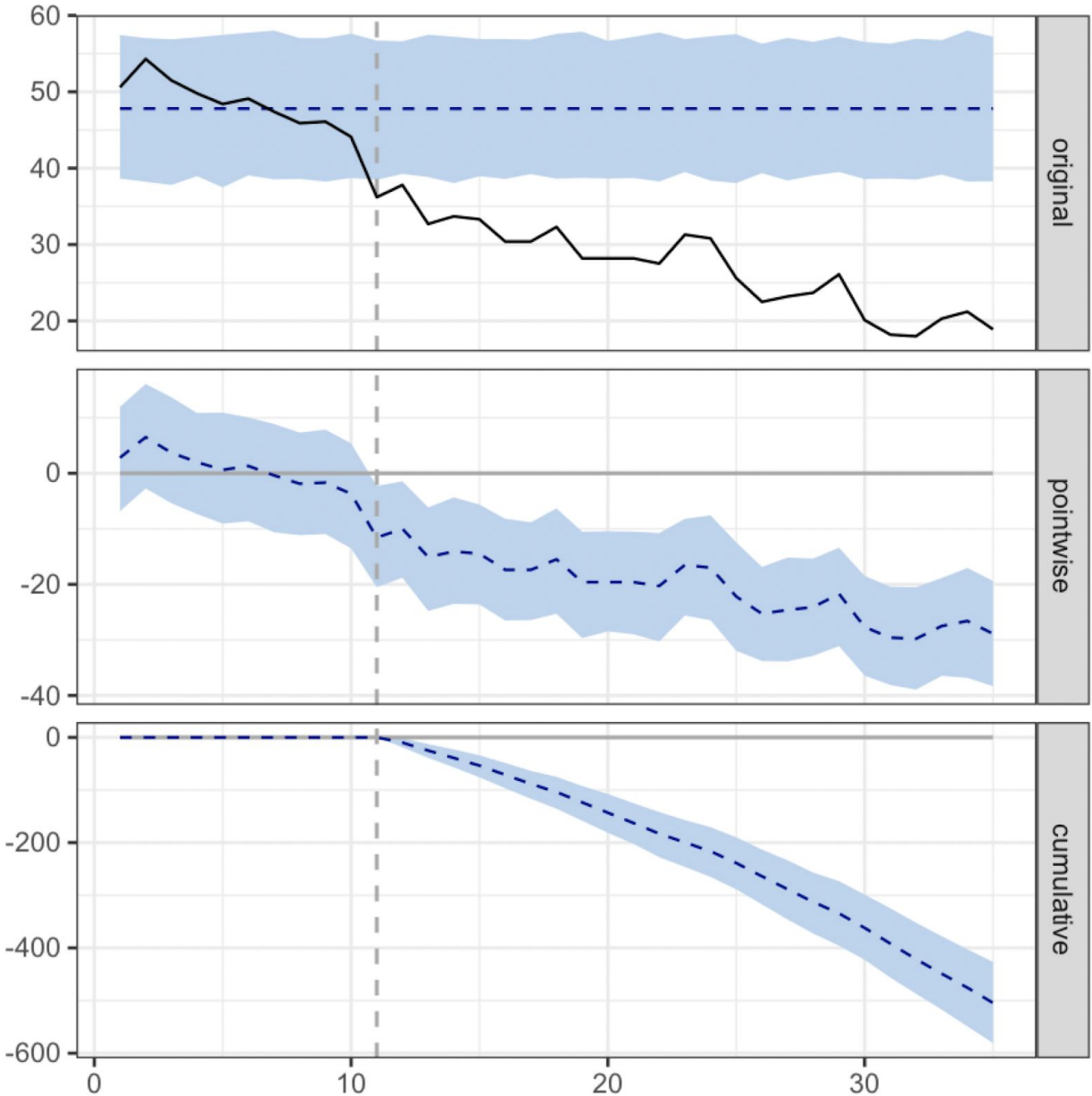

Figure 8: The X axis is the number of years passed since 1994.
Top panel: shows the counterfactual estimate (dashed line) and confidence intervals for it, and the solid line is the actual observed values (beyond 1994, the intervention event of hyperinflation completion). Middle panel: Shows the estimated treatment effect, here the difference in the forest cover yearly between no intervention (projected counterfactual) and actual observed forest cover values. E.g.: at X=25, which is the year 2019, the tree cover difference is -25% (reduction) as compared to the projected no-intervention or 'before' values, this shows that the end of hyperinflation event did cause an increased rate of deforestation and frontier expansion and hence, higher reduction in forest cover over the years, which validates our hypothesis.

## 4. Conclusion

In this paper, an algorithm for causal inference using satellite imagery has been proposed. Computer vision and deep learning techniques are used to detect forest tree cover. This is applied to satellite images over time to get a time series, to which changepoint detection is applied, to identify the time of intervention. Next, a Bayesian causal inference technique is used to fit a model to the time series observed before the intervention point and extrapolate/estimate the counterfactual beyond the intervention time point. This projection is then compared to the observed time series post intervention, and taking their difference gives the treatment effect of the intervention. We apply the above algorithm to forest tree cover in Rondonia, Brazil, pre and post the end of hyperinflation event (in 1994) is considered as the intervention, and we estimate that the forest cover rate reduced at a much higher rate after than before the event, due to rapid human deforestation and frontier expansion after the hyperinflation ended.